\relax
\documentclass[letterpaper]{article} 
\usepackage{aaai21}  
\usepackage{times}  
\usepackage{helvet} 
\usepackage{courier}  
\usepackage[hyphens]{url}  
\usepackage{graphicx} 
\usepackage{natbib}  
\usepackage{caption} 
\usepackage{subcaption}
\usepackage{multirow}
\usepackage{floatrow}
\usepackage{makecell}
\usepackage{amsmath}
\newcommand\cincludegraphics[2][]{\raisebox{-0.3\height}{\includegraphics[#1]{#2}}}
\usepackage{booktabs}       
\usepackage{array}
\DeclareMathOperator{\sign}{sign}

\title{Transferable Universal Adversarial Perturbations Using Generative Models}
\author{

    Atiye Sadat~Hashemi,
        Andreas~Bär,
        Saeed~Mozaffari,
        and~Tim~Fingscheidt\\
}

\affiliations{
	\textsuperscript{\rm 1} Faculty of Electrical and Computer Engineering, Semnan University, Iran\\
	\{atiye.hashemi, saeed.mozaffari\}@semnan.ac.ir\\[0.3em]
	\textsuperscript{\rm 2} Institute for Communications Technology, Technische Universität Braunschweig, Germany\\
	\{andreas.baer, t.fingscheidt\}@tu-bs.de
}

\begin{document}

\maketitle
\begin{abstract}
Deep neural networks tend to be vulnerable to adversarial perturbations, which by adding to a natural image can fool a respective model with high confidence. Recently, the existence of image-agnostic perturbations, also known as universal adversarial perturbations (UAPs), were discovered. However, existing UAPs still lack a sufficiently high fooling rate, when being applied to an unknown target model. In this paper, we propose a novel deep learning technique for generating more transferable UAPs. We utilize a perturbation generator and some given pretrained networks so-called source models to generate UAPs using the ImageNet dataset. Due to the similar feature representation of various model architectures in the first layer, we propose a loss formulation that focuses on the adversarial energy only in the respective first layer of the source models. This supports the transferability of our generated UAPs to any other target model. We further empirically analyze our generated UAPs and demonstrate that these perturbations generalize very well towards different target models. Surpassing the current state of the art in both, fooling rate and model-transferability, we can show the superiority of our proposed approach. Using our generated non-targeted UAPs, we obtain an average fooling rate of 93.36\% on the source models (state of the art: 82.16\%). Generating our UAPs on the deep ResNet-152, we obtain about a 12\% absolute fooling rate advantage vs. cutting-edge methods on VGG-16 and VGG-19 target models.
\end{abstract}

\section{Introduction}
Approaches relying on deep neural networks (DNNs) lead the benchmarks across several computer vision disciplines, including image classification \cite{ma2020autonomous}, object detection \cite{zhao2020gtnet}, and image segmentation \cite{poudel2019fast}.
Nonetheless, while showing superior performance on clean data, DNNs have been shown to be significantly vulnerable to small but maliciously structured perturbations to the input, known as \textit{adversarial perturbations}. Several hypotheses have been made to explain the existence of adversarial perturbations, such as poor regularization \cite{szegedy2013intriguing}, model linearity \cite{goodfellow2014explaining}, texture biased architectures \cite{geirhos2018imagenet}, the shortage of training data \cite{shamir2019simple}, and absence of well-generalizing features \cite{ilyas2019adversarial}. With regard to these hypotheses, different approaches for creating adversarial perturbations have been introduced \cite{ moosavi2016deepfool, zhao2019adversarial,zhao2020towards}.

Adversarial perturbations in computer vision tasks can be divided into two types, image-dependent perturbations and image-agnostic perturbations, the latter also known as \textit{universal adversarial perturbations} (UAPs). Image-dependent perturbations intrinsically depend on data samples and are usually estimated by solving an optimization problem \cite{bastani2016measuring} or using iterative/non-iterative gradient descent algorithms \cite{goodfellow2014explaining}. Consequently, finding an adversarial perturbation for a new image involves solving a new image-dependent optimization problem from scratch.
In return, UAPs are more generalizable perturbations that by adding to any image taken from a specific dataset lead to the deception of an underlying network in almost all cases.

Besides various algorithms for creating UAPs, generative models based on DNNs have also received more attention lately \cite{reddy2018ask, song2018constructing}. Some researchers proposed the use of generative adversarial networks (GANs) \cite{goodfellow2014generative} in combination with adversarial training to increase the robustness of a DNN to adversarial examples \cite{xiao2018generating}. However, combination of fooling and discriminative losses in typical GANs training, led to sub-optimal results in the case of generating adversarial perturbations. It persuaded researchers to train only a generative model through a single well behaved optimization to carry out adversarial attacks \cite{poursaeed2018generative,reddy2018nag}. In this paper, we also leverage training a generative model for producing UAPs.

In general, adversarial attacks can be categorized into white-box and black-box attacks. In the white-box setting, the parameters as well as the architecture of a model are accessible for a potential attacker. In the black-box setting, a potential attacker is neither able to access the model parameters nor the model architecture and thus has to rely on a good guess. It is of special interest to create adversarial examples that are able to not only fool one specific network, but also several other networks as well, that are trained on the same dataset. The ability of an adversarial example to be able to fool more than one network is often referred to as its \textit{transferability} \cite{papernot2016transferability}. Several approaches have been suggested for enhancing the transferability of black-box attacks \cite{wu2018understanding,li2020learning}. In this paper, we also aim at increasing the transferability of universal adversarial perturbations. Our contributions are as follows:

{\bf First}, we provide some analysis on the similarity of extracted feature maps, from the first activation layer in various different state-of-the-art architectures, using the structural similarity (SSIM) index.

{\bf Second}, in consequence, we propose a new loss function in which the fast feature fool loss \cite{mopuri2018generalizable}, focusing on the first layer only, is combined with the cross-entropy loss to train a generator using a source model. The aim is to generate UAPs with a high model transferability.

{\bf Finally}, we conduct targeted and non-targeted attacks on the ImageNet \cite{russakovsky2015imagenet} dataset, showing the effectivity of our proposed approach in terms of fooling rate and model transferability. When compared to other data-driven and data-independent attacks, our proposed method achieves the highest fooling rate as well as a better transferability across different models on the ImageNet dataset.

\section{Background}
The field of adversarial deep learning investigates different approaches for attacking on networks and defending against adversarial attacks \cite{jan2019connecting}. In general, adversarial attacks aim at perturbing clean data by adding an adversarial perturbation to it. In this section, we introduce our mathematical notations, two general types of adversarial perturbations, as well as the concept of transferability in this field.

\subsection {Basic Mathematical Notations}\label{section.notation}
Let $\mathcal{T}$ be the target model under attack, which is a deep neural network with frozen parameters, pretrained on an image dataset $\mathcal{X}^{\textrm{train}}$ and inferred on another dataset 
$\mathcal{X}^{\textrm{test}}$. In addition, we define the source model $\mathcal{S}$ as a pretrained model for which an adversarial perturbation $\boldsymbol{r}$ is generated with the use of a generator model $\mathcal{G}$. We define $\boldsymbol{z}$ as a random variable sampled from a distribution, which is fed to the generator $\mathcal{G}$ to produce a perturbation $\boldsymbol{r}=\mathcal{G}(\boldsymbol{z})$. Let $\boldsymbol{x}\in[0,1]^{H\times W\times C}$ be a normalized clean image with height $H$, width $W$, and $C=3$ color channels as dimensions, taken from any clean image set $\mathcal{X}^\textrm{train}$ or $\mathcal{X}^\textrm{test}$. Each image $\boldsymbol{x}$ is tagged with a ground truth label $ \overline{m} \in \mathcal{M}=\{1,2,...,M\}$. We define $\boldsymbol{y}$ as the network prediction for the input image $\boldsymbol{x}$, i.e., in training phase $\boldsymbol{y}=\mathcal{S}(\boldsymbol{x})$ and in test phase $\boldsymbol{y}=\mathcal{T}(\boldsymbol{x})$, with the output vector $\boldsymbol{y}=(y_\mu)$ and $m=\arg \underset{\mu \in \mathcal{M}}\max \ y_\mu$. Let $\mathcal{X}_{\mathcal{S}}^{\textrm{adv}}$ denote the adversarial space for the model $\mathcal{S}$, i.e., $\boldsymbol{x}^\textrm{adv} \in \mathcal{X}^\textrm{adv}_\mathcal{S}$, where $\boldsymbol{x}^\textrm{adv} = \boldsymbol{x} + \boldsymbol{r}$ is an adversarial example. In a similar way, $\mathcal{X}^\textrm{adv}_{\mathcal{T}}$ represents the adversarial space of the model $\mathcal{T}$. When $\mathcal{S}(\boldsymbol{x}^\textrm{adv}) = (y_\mu)$, the desired network output in non-targeted attack is $m=\arg \underset{\mu \in \mathcal{M}}\max \ y_\mu \ne \overline{m}$, whereas in targeted attacks it is $ m=\mathring{m}\ne \overline{m}$, with target class $\mathring{m}$. In order to have a quasi-imperceptible perturbation when added to clean images, we define $\|\boldsymbol{r}\|_p\le\epsilon$, with $\epsilon$ being the supremum of a respective $p$-norm $\|\cdot \|_p$. Also, let $J$ stand for a loss function.

\subsection{Image-dependent Adversarial Perturbations} \label{section.dependent}
From the time researchers have demonstrated the existence of adversarial perturbations \cite{szegedy2013intriguing}, different attacks have been introduced to craft adversarial examples more effectively and efficiently \cite{zhao2019adversarial}. Most common attacks are gradient-based methods, where typically an already trained model is used to craft adversarial examples based on the gradient with respect to the input using a loss function. Goodfellow et al.~\shortcite{goodfellow2014explaining} introduced the fast gradient sign method (FGSM), which is one of the most popular adversarial attacks. FGSM is defined as
\begin{equation}
\boldsymbol{x}^{\textrm{adv}}= \boldsymbol{x} + \boldsymbol{r} = \boldsymbol{x}+\beta\cdot \sign (\boldsymbol{\nabla}_{\boldsymbol{x}}J(\mathcal{S}(\boldsymbol{x}),\overline{\boldsymbol{y}})),
\end{equation}
where $\beta$ is a hyperparameter controlling the infinity norm of the underlying adversarial example, $\boldsymbol{y}=\mathcal{S}(\boldsymbol{x})$ is the output of the source model $\mathcal {S}$ utilized for producing adversarial examples, $\overline{\boldsymbol{y}}=(\overline{y}_\mu)$ is the one-hot encoding of the ground truth label $\overline{m}$ for image $\boldsymbol{x}$, while $\boldsymbol{\nabla}_{\boldsymbol{x}}J(\cdot)$ are the gradients with respect to the input under the loss function $J(\cdot)$.
Iterative FGSM \cite{kurakin2016adversarial}, iteratively applies FGSM with a small step size, while momentum FGSM \cite{dong2018boosting} utilizes a momentum-based optimization algorithm for stronger adversarial attacks. Besides, Su et al.~\shortcite{su2019one} presented an algorithm that efficiently locates one pixel (or a small set of pixels) to be perturbed for creating an adversarial example, without using any gradient information. The drawback of this method is the high computational complexity due to the dependence on data pixels.

\begin{figure*}[t!]
\centering
\hspace{0cm}%
    \begin{subfigure}[b]{0.51\textwidth}
        \includegraphics[width=\textwidth]{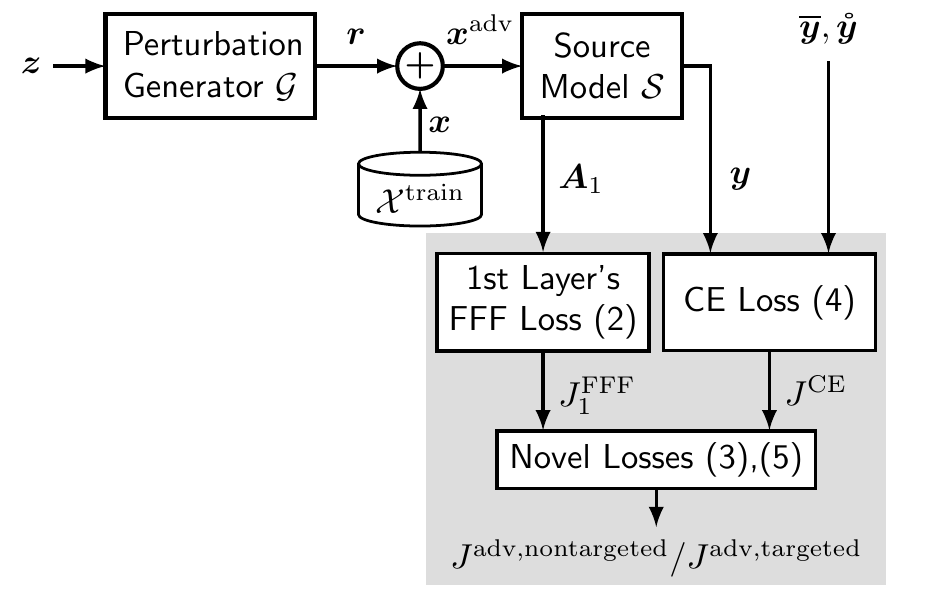} 
        \caption{Training phase}
        \label{Training phase}
    \end{subfigure}
    \begin{subfigure}[b]{0.46\textwidth}
        \includegraphics[width=\textwidth]{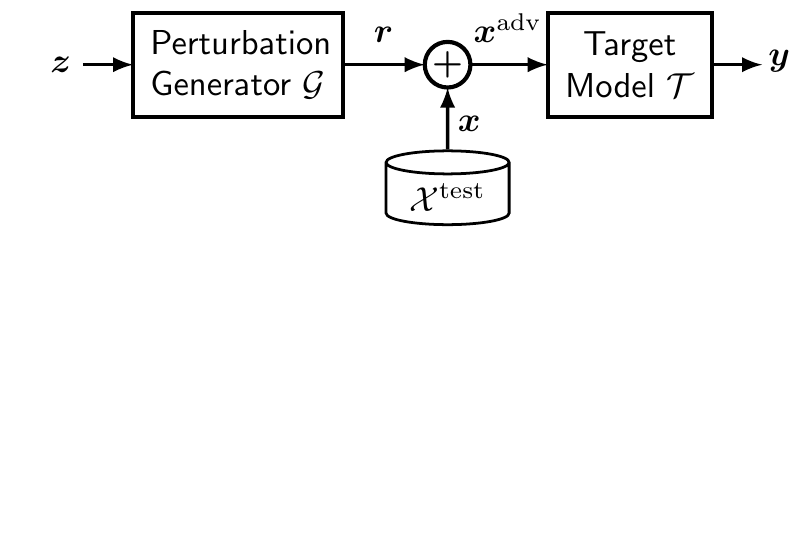} 
        \caption{Test phase}
        \label{Test phase}
    \end{subfigure}

\caption{Our proposed approach to generate UAPs for non-targeted and targeted attacks.}{\label{fig:our architecture}}
\end{figure*}

For improving the model robustness against adversarial attacks, researchers reported encouraging results by including adversarial examples during the training phase, known as adversarial training \cite{szegedy2013intriguing, tramer2019adversarial, dai2020adversarial}. Adv-GAN and Noise-GAN, networks which build upon GANs for generating and optimizing adversarial training, were also introduced \cite{xiao2018generating,hashemi2019secure}. Moreover, several methods including detecting adversarial attacks \cite{tian2018detecting,yang2020ml}, pre-processing approaches \cite{prakash2018deflecting, ding2019advertorch}, and utilizing teacher-student frameworks \cite{bar2019robustness} have been proposed for defending against attacks.
\subsection{Universal Adversarial Perturbations} \label{section.agnostic}
Image-agnostic perturbations, known as universal adversarial perturbations (UAPs), were firstly introduced by Moosavi-Dezfooli et al.~\shortcite{moosavi2017universal}. They proposed an iterative algorithm to generate UAPs to fool a classifier on a specific dataset. They also provided an analytical analysis of the decision boundary in DNNs based on geometry and proved the existence of small UAPs \cite{moosavi2018robustness}. Khrulkov et al.~\shortcite{khrulkov2018art} proposed to compute the singular vectors of the Jacobian matrices of a DNN's hidden layers to obtain UAPs. Hayes et al.~\shortcite{hayes2018learning} focused on generative models that can be trained for generating perturbations, while Poursaeed et al.~\shortcite{poursaeed2018generative} introduced the generative adversarial perturbation (GAP) algorithm for transforming random perturbations drawn from a uniform distribution to adversarial perturbations in order to conduct targeted and non-targeted attacks in classification and segmentation tasks. 

Contrary to previous papers, Mopuri et al.~\shortcite{mopuri2018generalizable} introduced fast feature fool (FFF), a \textit{data-independent} algorithm for producing non-targeted UAPs. In FFF, a new loss function is defined to inject maximal adversarial energy into each layer of a network as
\begin{equation}
\label{fff_formula}
\begin{split}
J^{\textrm{FFF}}(\boldsymbol{r})= \sum_{\ell=1}^{L}J_{\ell}^{\textrm{FFF}}(\boldsymbol{r})
\quad \text{with} \quad\\ J_{\ell}^{\textrm{FFF}}(\boldsymbol{r})=-\log(\|\boldsymbol{A}_{\ell}(\boldsymbol{r})\|_{2}),
\end{split}
\end{equation}
where $\boldsymbol{A}_{\ell}(\boldsymbol{r})$ is the mean of all feature maps of the $\ell$-th layer (after the activation function in layer $\ell$), when only the UAP $\boldsymbol{r}$ is fed into the model. The proposed FFF algorithm starts with a random $\boldsymbol{r}$ and is then iteratively optimized. For mitigating the absence of data in producing UAPs, Mopuri et al.~\shortcite{reddy2018ask} introduced class impressions (CIs), which are reconstructed images that are obtained via simple optimization from the source model. After finding multiple CIs in the input space for each target class, they trained a generator to create adversarial perturbations. By using this method, they managed to reduce the performance gap between the data-driven and data-independent approaches to craft UAPs. 

Several approaches have been proposed for defending against universal perturbations. Mummadi et al.~\cite{mummadi2019defending} have shown that adversarial training is surprisingly
effective in defending against UAPs. Some countermeasures define a distribution over such adversarial perturbations for a secure deep neural network. This can be done by learning a generative model \cite{hayes2018learning} or by finetuning model parameters to become more robust against this distribution of perturbations \cite{moosavi2018robustness}. These approaches are prone to overfit to a specific adversarial distribution, however, they increase model robustness against UAPs to some level. Recently Shafahi et al. \shortcite{shafahi2020universal} introduced universal adversarial training, which models the problem of robust model generation as a two-player min-max game, and produces robust classifiers. 
Also, some works including Akhtar et al.~\shortcite{akhtar2018defense} proposed a rectification and detection system against UAPs.

\subsection{Transferablity} \label{section.Transferablity}
The transferability of adversarial examples across different models has been studied experimentally and theoretically \cite{tramer2017space, phan2020cag}. Goodfellow et al.~\shortcite{goodfellow2014explaining} demonstrated that adversarial changes happen in large, contiguous areas in data rather than being thrown loosely into little image regions. Therefore, estimating the size of these adversarial subspaces is relevant to the transferability issue. Another perspective about transferability lies in the similarity of decision boundaries. Learning substitute models, approximating the decision boundaries of target models, is one famous approach to attack an unknown model \cite{papernot2016limitations}. Wu et al.~\shortcite{wu2020skip} considered neural networks with skip connections and found that using more gradients from the skip connections rather than the residual modules, allows the attacker to craft more transferable adversarial examples. Wei et al.~\shortcite{wei2018transferable} manipulate the feature maps extracted by a separate feature network, beside a generative adversarial network to create more transferable image-dependant perturbations. Also, Li et al.~\shortcite{li2020learning} introduced a virtual model known as Ghost network to apply feature-level perturbations to an existing model to create a large set of diverse models. They showed Ghost networks, together with the coupled ensemble strategy, improve the transferability of existing techniques. In addition, Wu et al.~\shortcite{wu2018understanding} empirically investigated the dependence of adversarial transferability to model-specific attributes, including model capacity, architecture, and test accuracy. They demonstrated that fooling rates heavily depend on the similarity of the source model and target model architectures. In this paper, we make use of the similarity of low-level extracted features in the initial layers of several models to improve the transferability of generated UAPs.

\begin{figure}[!t]
	\setlength\tabcolsep{0.4pt}
	\captionsetup{type=figure}
	\begin{tabular}
	{c c c c c c c}
		& $\ell=1$ & $\ell=2$ & $\ell=3$ & $\ell=4$ & $\ell=5$ & $\ell=6$\\
		
		\makecell{VGG\\16}
		&\makecell{\cincludegraphics[scale=0.14]{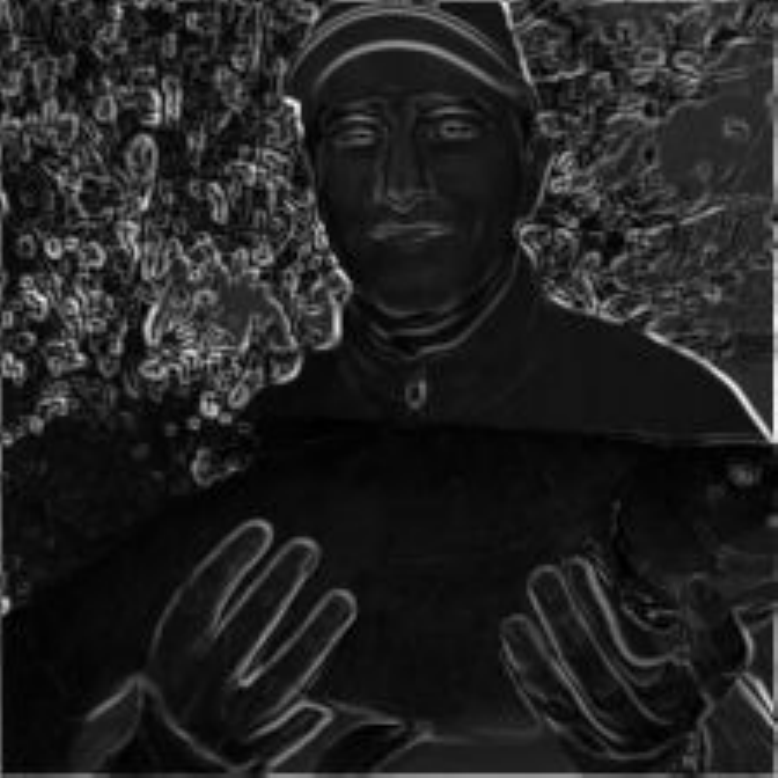}}
		&\makecell{\cincludegraphics[scale=0.14]{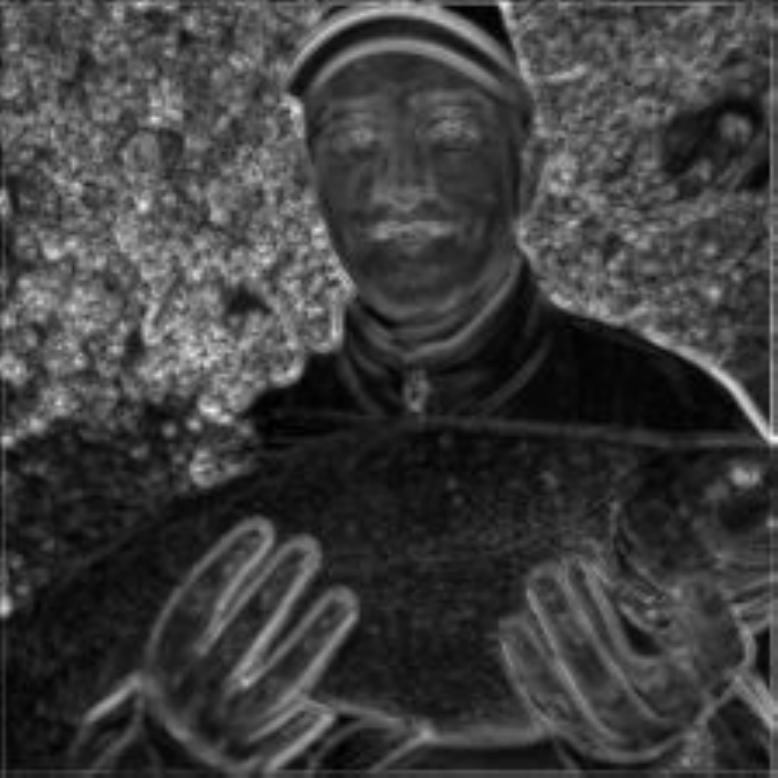}}
		&\makecell{\cincludegraphics[scale=0.28]{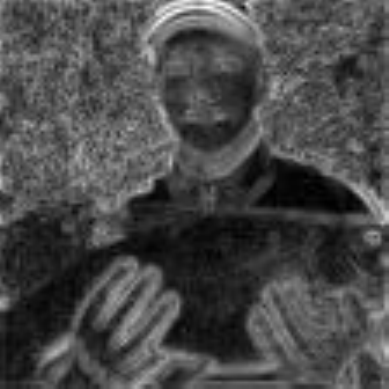}}
		&\makecell{\cincludegraphics[scale=0.28]{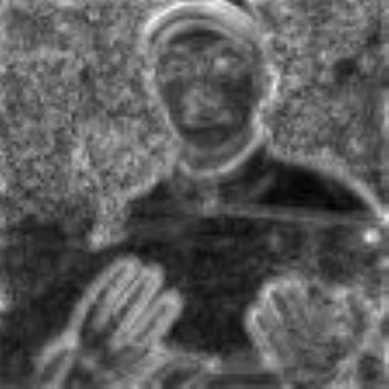}}
		&\makecell{\cincludegraphics[scale=0.56]{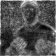}}
		&\makecell{\cincludegraphics[scale=0.56]{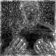}}\\[0.4cm]

		\makecell{VGG\\19}
		&\makecell{\cincludegraphics[scale=0.14]{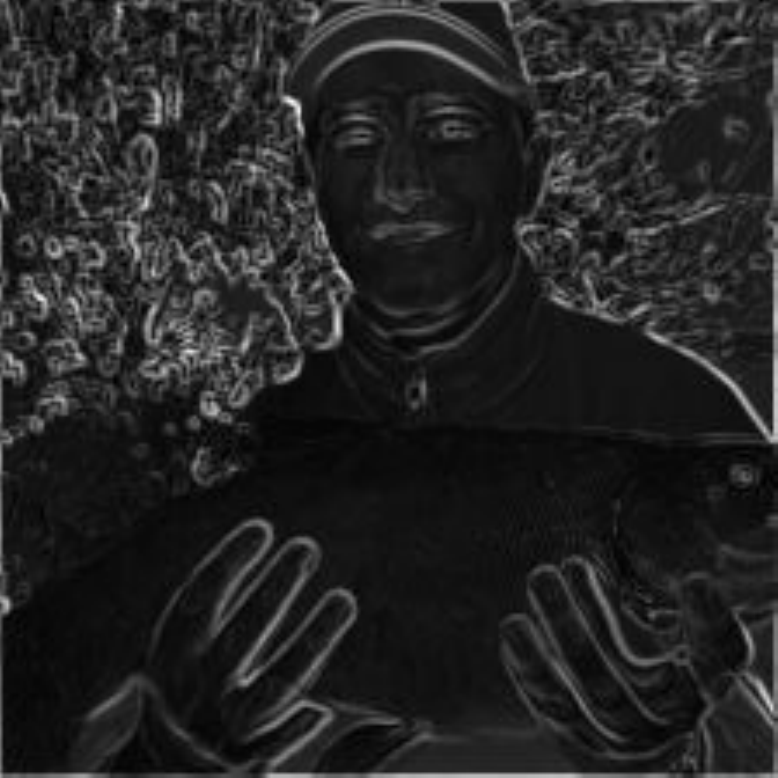}}
		&\makecell{\cincludegraphics[scale=0.14]{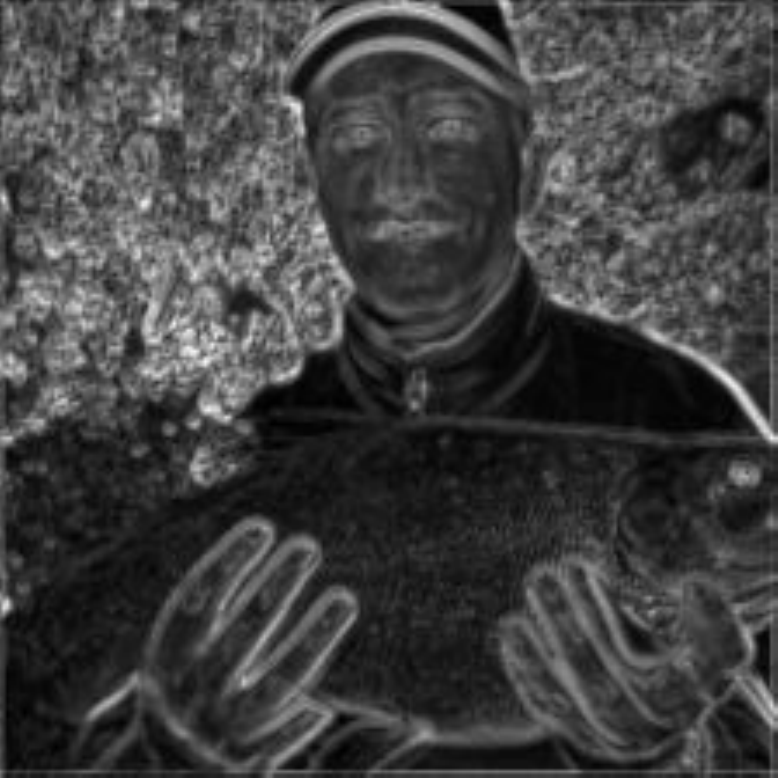}}
		&\makecell{\cincludegraphics[scale=0.28]{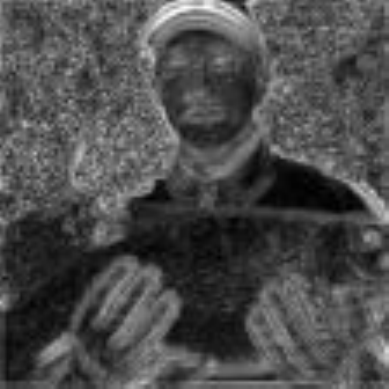}}
		&\makecell{\cincludegraphics[scale=0.28]{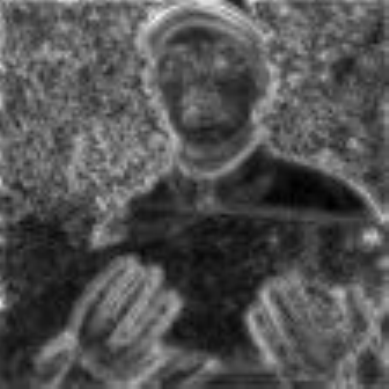}}
		&\makecell{\cincludegraphics[scale=0.56]{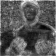}}
		&\makecell{\cincludegraphics[scale=0.56]{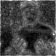}}\\[0.4cm]

		\makecell{ResNet\\18}
		&\makecell{\cincludegraphics[scale=0.28]{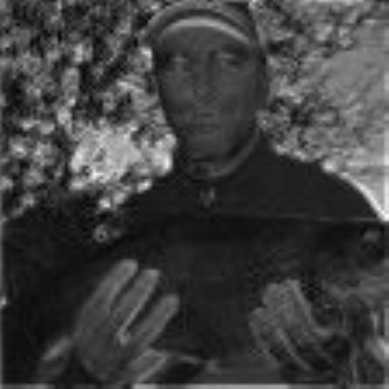}}
		&\makecell{\cincludegraphics[scale=0.56]{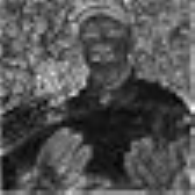}}
		&\makecell{\cincludegraphics[scale=0.56]{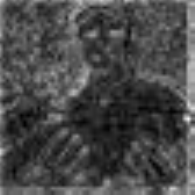}}
		&\makecell{\cincludegraphics[scale=1.13]{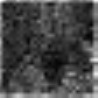}}
		&\makecell{\cincludegraphics[scale=1.13]{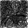}}
		&\makecell{\cincludegraphics[scale=2.26]{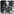}}\\[0.4cm]

		\makecell{ResNet\\152}
		&\makecell{\cincludegraphics[scale=0.28]{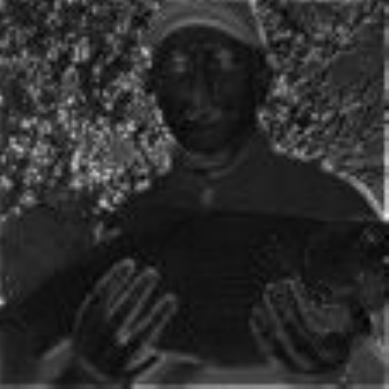}}
		&\makecell{\cincludegraphics[scale=0.56]{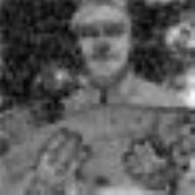}}
		&\makecell{\cincludegraphics[scale=0.56]{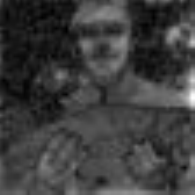}}
		&\makecell{\cincludegraphics[scale=0.56]{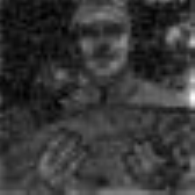}}
		&\makecell{\cincludegraphics[scale=1.13]{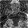}}
		&\makecell{\cincludegraphics[scale=1.13]{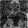}}\\[0.4cm]
	\end{tabular}
	\captionof{figure}{The layer-wise mean of feature representations $\boldsymbol{A}_{\ell}(\boldsymbol{x})$ within different pretrained classifiers, computed for the first six layers for an input image $\boldsymbol{x}$.}
	\label{fig:Feature maps}
\end{figure}

\section{Method} \label{section.Our method}
At the core of our method there is an adversarial perturbation generator which employs a pretrained network as the source model $\mathcal{S}$ to generate universal adversarial perturbations for conducting targeted and non-targeted attacks. Our proposed method builds upon the network proposed by Poursaeed et al.~\shortcite{poursaeed2018generative}. Our goal is to find a perturbation $\boldsymbol{r}$ which is able to not only fool the source model $\mathcal{S}$ on a set of images $\mathcal{X}^\text{train}$ or $\mathcal{X}^\text{test}$, respectively, but the perturbation $\boldsymbol{r}$ should be also effective on target models $\mathcal{T}$, for which $\mathcal{T}\ne\mathcal{S}$ holds. 

\subsection{Generating Universal Adversarial Perturbations} \label{section.details1}
Figure~\ref{fig:our architecture} illustrates the architecture of our model for generating UAPs. A multi-dimensional input $\boldsymbol{z}\in [0, 1]^{H\times W\times C}$ sampled from a uniform distribution is fed to the perturbation generator $\mathcal{G}$. The network $\mathcal{G}$ outputs the adversarial perturbation $\boldsymbol{r}$, which is bounded according to a $p$-norm. We follow \cite{poursaeed2018generative} and bound the perturbation by multiplying the generator network raw output $\mathcal{G}'(\boldsymbol{z})$ with $\min(1,\frac{\epsilon}{\|\mathcal{G}'(\boldsymbol{z})\|_p})$ to obtain a scaled UAP $\boldsymbol{r}$. In the training phase, the resulting adversarial perturbation $\boldsymbol{r}$ is added to a natural image $\boldsymbol{x}\in \mathcal{X}^{\textrm{train}}$ and clipped to a valid range of RGB image pixel values to create an adversarial example $\boldsymbol{x}^{\textrm{adv}}$. The generated adversarial example is then fed to a source model $\mathcal{S}$ to compute the cross-entropy loss $J^\textrm{CE}$ as well as the first layer's fast feature fool loss $J^\textrm{FFF}_1$, see (\ref{fff_formula}). For training the network $\mathcal{G}$, depending on targeted or non-targeted attacks, the process differs as follows.
\paragraph{Non-targeted Perturbations} In this case, we want to fool the network $\mathcal{S}$ so that its prediction $m$ differs from the ground truth $\overline{m}$. In the simplest possible way, we can define the negative cross-entropy as the fooling loss for non-targeted attacks. To increase the transferability of the produced UAPs across models, we seek for similarities between different pretrained models. We selected VGG-16, VGG-19 \cite{simonyan2014very}, ResNet-18, and ResNet-152 \cite{he2016deep}, all pretrained on ImageNet \cite{russakovsky2015imagenet}, as state-of-the-art DNN classifiers to explore their learned feature maps. Figure~\ref{fig:Feature maps} shows the mean of feature representations $\boldsymbol{A}_{\ell}(\boldsymbol{x})$ of these pretrained classifiers computed for layer $\ell=1$ up to $\ell=6$ (after activation function) for a randomly selected input image $\boldsymbol{x}$. Also, Table~\ref{tab:Feature maps2} shows the similarity between these mean feature maps $\boldsymbol{A}_{\ell}(\boldsymbol{x})$ in terms of structural similarity (SSIM) \cite{wang2004image} as the evaluation criteria. SSIM is applied to measure the perceptual difference between low and high level feature maps, where the ImageNet validation set \cite{russakovsky2015imagenet} has been used as networks inputs. As this analysis shows, the mean of the extracted feature maps in the first layers of these classifiers are more similar to each other and the deeper they get, the less similar they become. We thus hypothesize that by applying the fast feature fool loss only to the first layer of the source model, with the aim of injecting high adversarial energy into the first layer of the source model $\mathcal{S}$ during training of the perturbation generator $\mathcal{G}$, the transferability of generated UAPs increases. Then, we define the generator fooling loss for our non-targeted attacks as

\begin{table}[t]
	\captionof{table}{Structural similarity (SSIM) index between the mean feature representations of VGG-16 \cite{simonyan2014very} and different classifiers pretrained on ImageNet (see Figure \ref{fig:Feature maps}) in layers $\ell\in\lbrace 1,2,3,4,5,6 \rbrace$. The ImageNet validation set has been used in this experiment. All networks show a considerable structural similarity in the first layer, and only VGG-19 \cite{simonyan2014very} in the later layers. The \mbox{highest} SSIM for each network is printed in \textbf{boldface}.}
	\setlength\tabcolsep{4.9pt}
\centering
	\begin{tabular*}{0.95\textwidth}{c| c c c c c c}
		\toprule
		&\multicolumn{6}{c}{The number of layer ($\ell$)}\\
		& $1$ & $2$  & $3$ & $4$ & $5$ & $6$ \\
		\midrule
		VGG-19& \bf{0.94} & 0.85 & 0.77 & 0.72&0.77&0.60 \\
		ResNet-18 & \bf{0.58} & 0.29 & 0.15 & 0.06& 0.01& 0.00\\
		ResNet-152& \bf{0.57} & 0.17 & 0.07 & 0.05& 0.01& 0.01\\
		\bottomrule
	\end{tabular*}
	\label{tab:Feature maps2}
\end{table}

\begin{equation}
J^{\textrm{adv},\textrm{nontargeted}}=\alpha\cdot(- J^{\textrm{CE}}(\mathcal{S}(\boldsymbol{x}^{\textrm{adv}}),\boldsymbol{\overline{y}}))+ (1-\alpha)\cdot J^{\textrm{FFF}}_{1}(\boldsymbol{x}^{\textrm{adv}}),
\label{formula4}
\end{equation}
where $J^{\textrm{CE}}$ denotes the cross-entropy loss, and $\boldsymbol{\overline{y}}=(\overline{y}_\mu)$ is the one-hot encoding of the ground truth label $\overline{m}$ for image $\boldsymbol{x}$, and $\mu$ being the class index. Also, $J^{\textrm{FFF}}_{1}(\boldsymbol{x}^{\textrm{adv}})$ is the fast feature fool loss of layer $\ell=1$, when $\boldsymbol{x}^{\textrm{adv}}$ is fed to the network $\mathcal{S}$ resulting in $\boldsymbol{y}= \mathcal{S}(\boldsymbol{x}^{\textrm{adv}})$. The cross-entropy loss is obtained by
\vspace{0.2cm}
\begin{equation}
J^{\textrm{CE}}(\boldsymbol{y}, \boldsymbol{\overline{y}}) = - \sum_{\mu \in \mathcal{M}} \overline{y}_\mu \log \left(y_{\mu}\right),
\end{equation}
where $\boldsymbol{y} = ({y}_\mu)$ is the output vector of the network $\mathcal{S}$ with the predictions for each class $\mu$. Then, we utilize the Adam optimizer \cite{kingma2014adam} to increase the loss through a stochastic optimization.
\paragraph{Targeted Perturbations} Unlike to non-targeted attacks, the goal of a targeted one is $\mathcal{S}(\boldsymbol{x}^{\textrm{adv}})=(y_{\mu})$ with $\mathring{m}=\arg \underset{\mu \in \mathcal{M}} \max \ y_{\mu}$ and $\mathring{m}\neq \overline{m}$, where $\mathring{m}$ is the adversarial target label to be outputted by the attacked DNN, while $\overline{m}$ still denotes the ground truth. Hence, the attacker aims to decrease the cross-entropy loss with respect to a target $\mathring{m}$ until the source model $\mathcal{S}$ predicts the selected target class with high confidence. Also, we add the fast feature fool loss in the first layer to boost the transferability of the targeted generated UAP, resulting in our generator fooling loss for targeted attacks as
\vspace{0.2cm}
\begin{equation}
J^{\textrm{adv,targeted}}=\alpha\cdot J^{\textrm{CE}}(S(\boldsymbol{x}^{\textrm{adv}}),\mathring{\boldsymbol{y}})+(1-\alpha)\cdot J^{\textrm{FFF}}_{1}(\boldsymbol{x}^{\textrm{adv}}),
\label{formula5}
\end{equation}
where $\mathring{\boldsymbol{y}}$ is the one-hot encoding of the target label $\mathring{m}\neq \overline{m}$.

\begin{table}[]
	\centering
	\caption{
		Tuning the hyperparameter $\alpha$ in our non-targeted attack. The adversarial perturbation is bounded by $L_{\infty}(\boldsymbol{r})\le \epsilon = 10$. Results are reported on a \textbf{2nd training set}. The best fooling rates (\%) are printed in \textbf{boldface}.
		}
	\renewcommand{\arraystretch}{0.95} 
	\begin{tabular}{c| c c c c c}
		\toprule 
		  \multirow{3}{*}{$\alpha$} & \multicolumn{4}{c}{Source Model $\mathcal{S}$ = Target Model$\mathcal{T}$}  & \multirow{3}{*}{Avg}\\[0.25cm]
		  &VGG & VGG & ResNet & ResNet  \\
		  &16  & 19 & 18  & 152  \\
		\midrule
		 0  & 8.52 & 8.29  & 7.24 & 4.04 & 7.02\\
		 0.6  & 90.49 & 93.48 & 88.93 & 84.41 & 89.32\\
		 0.7  & \bf 95.20 & \bf 93.79 & \bf 89.16 &  87.05 & \textbf{91.30}\\
		 0.8  & 90.03 & 93.24 & 89.07 & \bf 89.91 & 90.56\\
		 0.9 & 95.13 & 92.14 & 88.34 &  89.37 & 91.24\\
		 1  & 92.87 & 71.88 & 88.88 & 85.34 & 84.74 \\
		\bottomrule
	\end{tabular}
	\label{tab:finetuning2}
\end{table}

\begin{figure*}[t!]
	\centering
	\begin{subfigure}[b]{\textwidth}
		\centering
		\includegraphics[width=0.088\linewidth]{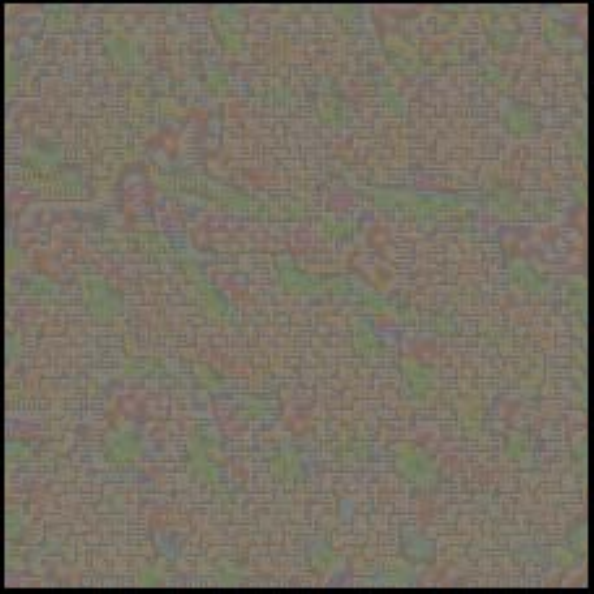}%
		\quad
		\includegraphics[width=0.7\linewidth]{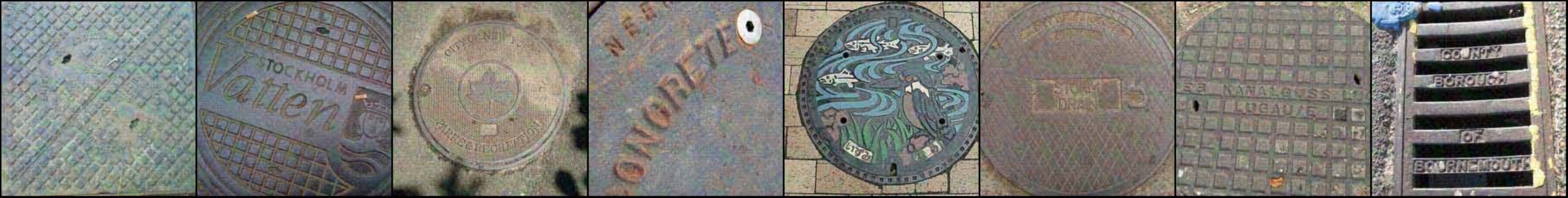}
		\caption{The UAP (left) and the respective adversarial examples (right), with $L_2(\boldsymbol{r})\le2000$.
		}
	\end{subfigure}
	\vskip\baselineskip
	\begin{subfigure}[b]{\textwidth}
        \hspace{35.5 mm}
        \includegraphics[width=0.7\linewidth]{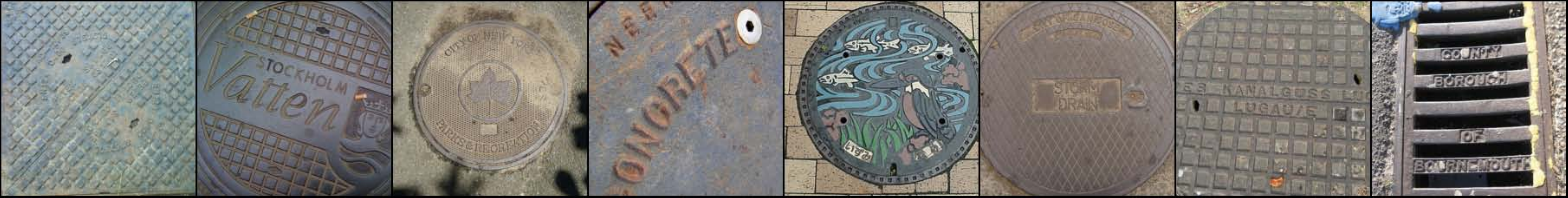}
		\caption{Original images.
		}
	\end{subfigure}
		\vskip\baselineskip
	\begin{subfigure}[b]{\textwidth}
		\centering
		\includegraphics[width=0.088\linewidth]{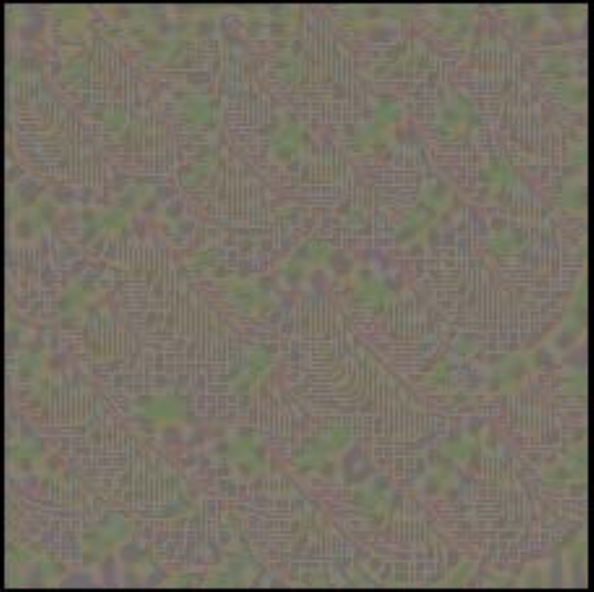}%
		\quad
		\includegraphics[width=0.7\linewidth]{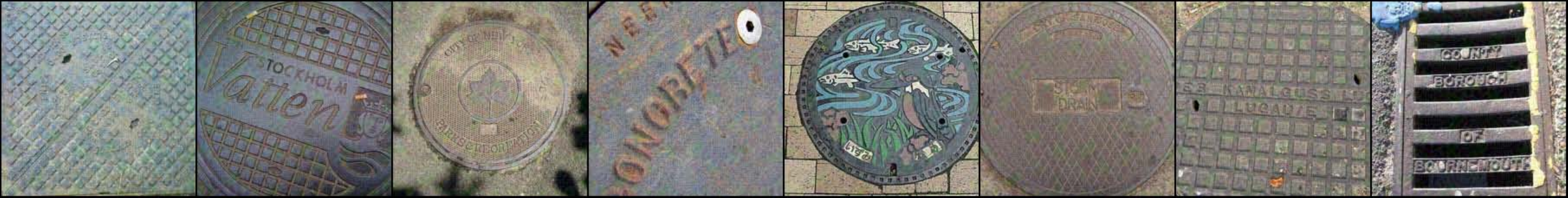}
		\caption{The UAP (left) and the respective adversarial examples (right), with $L_{\infty}(\boldsymbol{r})\le10$.
		}
	\end{subfigure}
	\caption{Examples of {\bf our non-targeted UAPs and adversarial examples}. In (a) the universal adversarial perturbation is given on the left and 8 different adversarial examples are shown on the right, where the $L_{2}$ norm of the adversarial perturbation is bounded by $\epsilon=2000$, i.e., $L_{2}(\boldsymbol{r})\le 2000$; in (b) the respective original images are shown, whereas in (c) the $L_{\infty}$ norm of the adversarial perturbation is bounded by $\epsilon=10$, i.e., $L_{\infty}(\boldsymbol{r})\le 10$, $\alpha=0.7$. 
	In these experiments, both the source model $\mathcal S$ and the target model $\mathcal T$ are VGG-16 \cite{simonyan2014very}.}
	\label{NON-TARGETED}
\end{figure*}

\section{Experimental Results} \label{section.Experiments and Results}
In this section, we present our experimental setup and analyze the effectiveness of our fooling method on state-of-the-art classifiers trained on ImageNet. In particular, we used VGG-16, VGG-19 \cite{simonyan2014very}, ResNet-18, and ResNet-152 \cite{he2016deep} as target classifiers $\mathcal{T}$. For all our experiments, a UAP is computed for a set of 10,000 images taken from the ImageNet training set (i.e., 10 images per class) and the results are reported on the ImageNet validation set (50,000 images).

There are several design options regarding the architecture choices for the generator $\mathcal{G}$ and the source model $\mathcal{S}$. For our generator, we follow \cite{zhu2017unpaired} and \cite{poursaeed2018generative} and choose the ResNet generator from \cite{johnson2016perceptual}, which consists of some convolution layers for downsampling, followed by some residual blocks before performing upsampling using transposed convolutions. In the case of the source model $\mathcal{S}$, we utilize the same pretrained classifiers as for the target model $\mathcal{T}$, i.e., VGG-16, VGG-19, ResNet-18, and ResNet-152. 
\subsection{Non-targeted Universal Perturbations} \label{section.details11}
In this section, we consider the target model $\mathcal{T}$ to be equal to the source model $\mathcal{S}$ used for optimizing the generator $\mathcal{G}$, which we refer to as the white-box setting. Similar to existing approaches \cite{moosavi2017universal, poursaeed2018generative, mopuri2018generalizable,reddy2018ask}, we use the \textit{fooling rate} as our metric to assess the performance of our crafted UAPs. In the case of non-targeted attacks, it is the percentage of input images for which $\mathcal{T}(\boldsymbol{x}^{\textrm{adv}})\ne \mathcal{T}(\boldsymbol{x})$ holds. According to Figure~\ref{fig:our architecture}, we train our model with the non-targeted adversarial loss function (\ref{formula4}). For tuning the hyperparameter $\alpha$, the weight of our novel adversarial loss components, we utilized another set of 10,000 images taken from the ImageNet training set which is different from our training dataset. Table \ref{tab:finetuning2} shows that the best $\alpha$ for non-targeted attacks, on average over all model topologies, is $\alpha=0.7$. Results on the ImageNet validation set for two different norms are given in Table~\ref{table1}. The maximum permissible $L_p$ norm of the perturbations for $p=2$ and $p=\infty$ is set to be $\epsilon = 2000$ and $\epsilon = 10$, respectively. As authors in 
\cite{moosavi2017universal} pointed out, these values are selected to acquire a perturbation whose norm is remarkably smaller than the average image norms in the ImageNet dataset to obtain quasi-imperceptible adversarial examples. The results in Table~\ref{table1} show that the proposed method is successful in the white-box setting. All reported fooling rate numbers are above 90\%. To illustrate that our adversarial examples are quasi-imperceptible to humans, we show some visual examples of generated UAPs as well as the adversarial and original images
in Figure~\ref{NON-TARGETED}.
\subsection{Targeted Universal Perturbations} \label{section.details22}
\begin{figure*}[!t]
	\centering
	\begin{subfigure}[b]{\textwidth}
		\centering{}
        \includegraphics[width=0.4\linewidth]{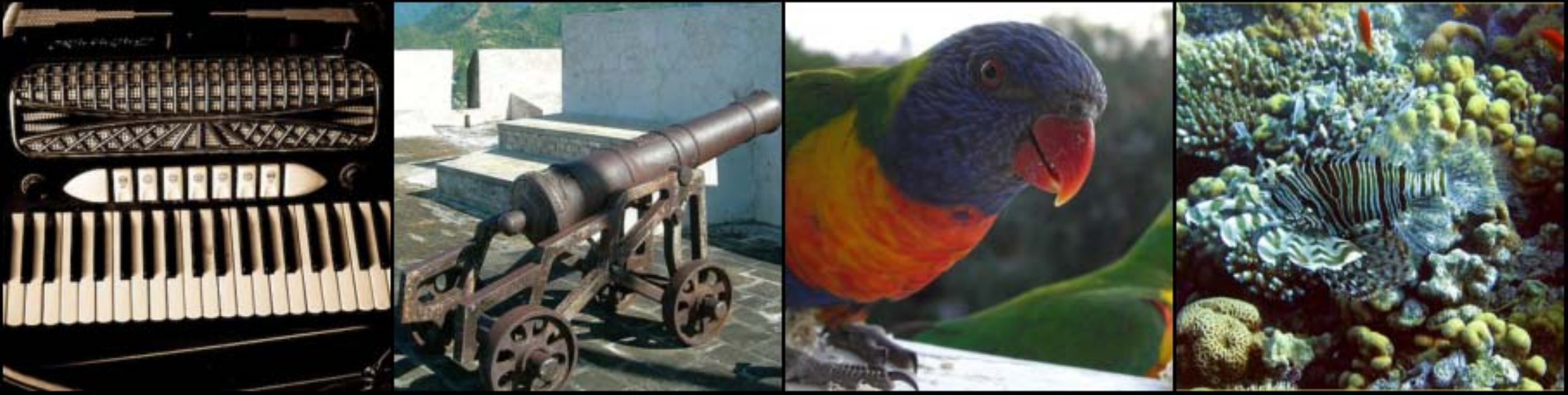}\quad
		\includegraphics[width=0.1\linewidth]{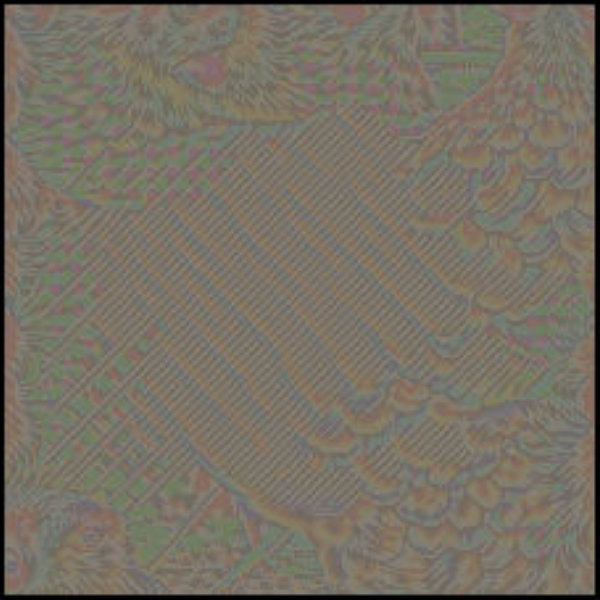}\quad
		\includegraphics[width=0.4\linewidth]{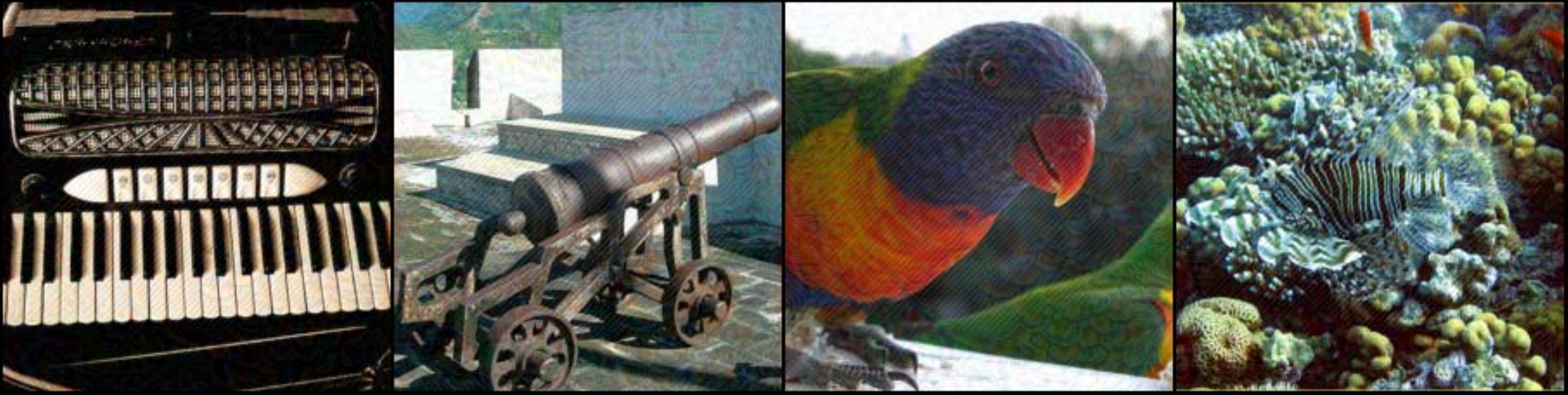}
	    \caption{Original images (left), UAP for target label "hen" (center), adversarial examples (right).
		}
	\end{subfigure}
		\vskip\baselineskip
	\begin{subfigure}[b]{\textwidth}
		\centering
        \includegraphics[width=0.4\linewidth]{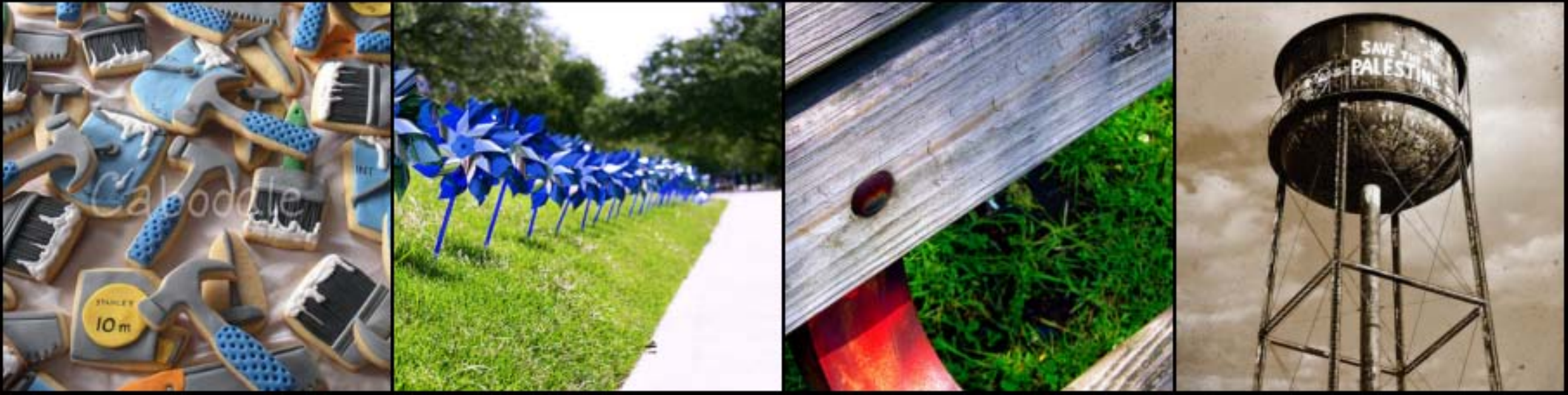}\quad
		\includegraphics[width=0.1\linewidth]{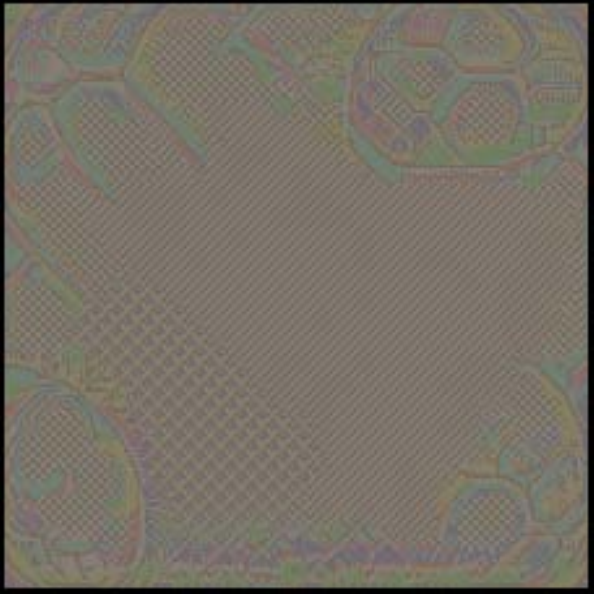}\quad
		\includegraphics[width=0.4\linewidth]{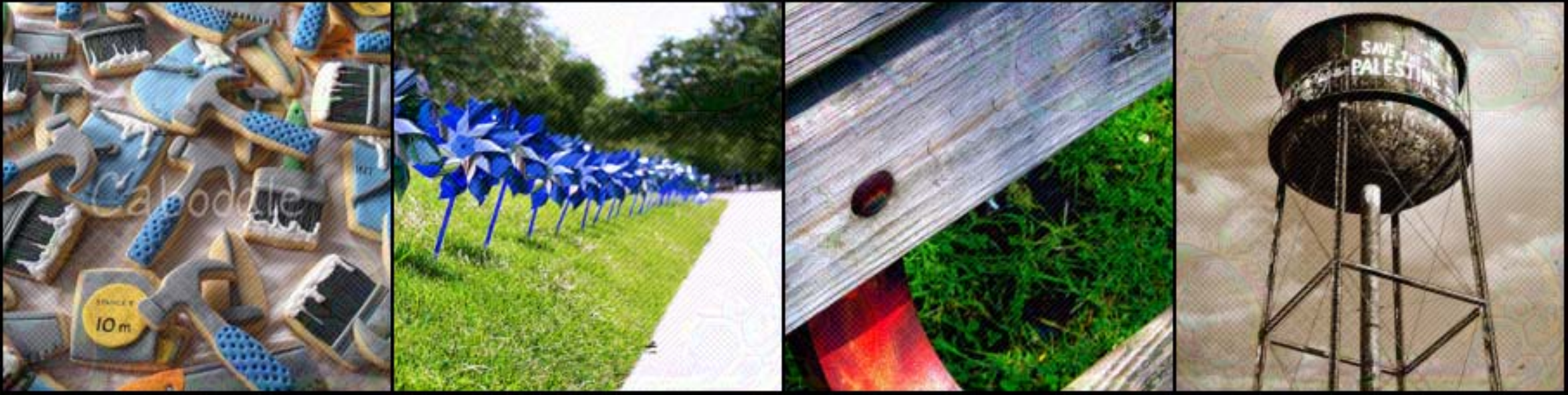}
		\caption{Original images (left), UAP for target label "soccer ball" (center), adversarial examples (right).
		}
	\end{subfigure}
	\caption{Examples of {\bf our targeted UAPs and adversarial examples}. Four different samples of original images are given on the left, the targeted UAP is shown in the middle and their respective adversarial examples are shown on the right. The target label in (a) is "hen", and in (b) is "soccer ball". In these experiments, both the source model $\mathcal S$ and the target model $\mathcal T$ are VGG-16 \cite{simonyan2014very}, with $L_{\infty}(\boldsymbol{r})\le10$, $\alpha=0.7$.
	}
	\label{fig:targeted}
\end{figure*}

In this section, we applied the targeted fooling loss function defined in (\ref{formula5}) again with $\alpha=0.7$ for training the generator in Figure~\ref{fig:our architecture}. In targeted attacks, we calculate \textit{top-1 target accuracy}, the ratio of adversarial examples which are classified as the desired target, as the attack success rate. Figure~\ref{fig:targeted} depicts two examples of our targeted UAPs, some original images and respective adversarial examples. In these experiments, the top-1 target accuracy on the validation set for the target class $\mathring{m}=8$ ("hen") and $\mathring{m}=805$ ("soccer ball"), are 74.65\% and 76.75\%, respectively, which underlines the effectiveness of our approach. 
Also, For assessing the generalization power of our proposed method across different target classes and comparison with targeted UAPs generated by the state-of-the-art GAP \cite{poursaeed2018generative}, we used 10 randomly sampled classes. The resulting average top-1 target accuracy, when the adversarial perturbation is bounded by $L_{\infty}(\boldsymbol{r})\le \epsilon = 10$, is 66.57\%, which is a significantly higher number than 52.0\%, that was reported for GAP \cite{poursaeed2018generative}.

\begin{table}[t]
	\caption{Fooling rates (\%) of {\bf our proposed non-targeted UAPs} for various target classifiers pretrained on ImageNet. Results are reported on the {\bf ImageNet validation set}. In these experiments, the source model $\mathcal{S}$ is the same as the target model $\mathcal{T}$. We report results for two $L_p$ norms, namely $L_2(\boldsymbol{r})\le \epsilon= 2000$ and $L_{\infty}(\boldsymbol{r}) \le\epsilon= 10$.}
	\label{table1}
	\centering
	\renewcommand{\arraystretch}{0.95} 
    \begin{tabular}{ccc|cccc}
     	\toprule
		\multirow{3}{*}{$p$} & \multirow{3}{*} {$\epsilon$} & \multirow{3}{*}{$\alpha$} & \multicolumn{4}{c}{Source Model $\mathcal S$ = Target Model $\mathcal T$}\\[0.25cm]
		& & & VGG& VGG & ResNet & ResNet \\
		& & & 16 & 19 & 18 & 152 \\
		\midrule
		2 & 2000 & 0.7 & 96.57 & 94.99 & 91.85 & 88.73  \\
		\midrule
		$\infty$ & 10 & 0.7 & 95.70 & 94.00 & 90.46 & 90.40\\
		\bottomrule
\end{tabular}
\end{table}

\begin{table}[t!]
	\centering
		\caption{Fooling rates (\%) of {\bf various non-targeted UAP methods} on various target classifiers trained on ImageNet. Our method is compared with other state-of-the-art methods. In these experiments, the source model is the same as the target model. The result of other attacks are reported from the respective paper. $^+$For comparison reasons, the average of our method leaves out the ResNet-18 model results. Best results are printed in \textbf{boldface}.}

    \setlength\tabcolsep{3pt} 
    \renewcommand{\arraystretch}{1} 
	\begin{tabular}{ccc|ccccc}
		\toprule
		\multirow{3}{*}{$p$} & \multirow{3}{*} {$\epsilon$} & \multirow{3}{*}{Method} & \multicolumn{4}{c}{$\mathcal{S}$ = $\mathcal{T}$} & \multirow{3}{*}{Avg$^+$}\\[0.2cm]
		& & & VGG & VGG & ResNet & ResNet & \\
		& & & 16 & 19 & 18 & 152 & \\
		\midrule
		\multirow{5}{*}{$\infty$} & \multirow{5}{*}{10} & 
		FFF & 47.10 & 43.62 & - & 29.78 & 40.16 \\
		& & CIs  & 71.59 & 72.84 & - & 60.72 & 68.38 \\
		& &UAP  & 78.30 & 77.80 & - & 84.00 & 80.03\\
		& & GAP  & 83.70 & 80.10 & - & 82.70 & 82.16\\
		&& {\bf Ours} & \bf 95.70 & \bf 94.00 & \bf 90.46 & \bf{90.40} & \bf 93.36\\
		\midrule
		\multirow{3}{*}{2} & \multirow{3}{*}{2000} & UAP & 90.30 & 84.50 & - & 88.50 & 87.76 \\ 
	     && GAP & 93.90  & 94.90 & - & 79.50 & 89.43\\ 
	     && {\bf Ours} & \bf 96.57 & \bf 94.99 & \bf 91.85 & \bf 88.73 & \bf 93.43\\ 
		\bottomrule
	\end{tabular}
	\label{tab:comparison}
\end{table}
\subsection{Transferability of Non-targeted UAPs} \label{section.details33}
To further investigate the performance of our generated UAPs, we analyze their transferability across different models. For this purpose, we craft a universal adversarial perturbation using the source model $\mathcal{S}$, and feed it into a target model $\mathcal{T}\ne\mathcal{S}$. Table~\ref{transferability} presents fooling rates for the proposed UAPs crafted for multiple pretrained models $\mathcal{S}$, across four different classifiers $\mathcal{T}$. For each source architecture $\mathcal{S}$ (first column), we compute a UAP and report the fooling rates on the same (main diagonal) and on all other networks $\mathcal{T}$. It can be observed that the proposed non-targeted UAPs are generalizing  very well across different architectures. The perturbation computed for ResNet-152 (as the source model $\mathcal{S}$), has an average fooling rate of 81.53\% on all other target models in Table~\ref{transferability}. 

\begin{table*}[t]
	\caption{{\bf Transferability of our proposed non-targeted UAPs}. The percentages indicate the fooling rates (\%). The generator is trained to fool the source model (rows), and it is tested on the target model (columns). The adversarial perturbation is bounded by $L_{\infty}(\boldsymbol{r})\le \epsilon = 10$, $\alpha=0.7$. $^+$ The average is computed without the white-box attacks (main diagonal).
	}
	\label{transferability}
	\centering
	\renewcommand{\arraystretch}{0.95} 
     \begin{tabular}{cc|ccccc}
     	\toprule
     	&&\multicolumn{4}{c}{Target model $\mathcal{T}$} & \multirow{2}{*}{Avg$^+$}\\[0.05cm]
		&&VGG-16 & VGG-19 & ResNet-18 & ResNet-152 \\
	    \midrule
	    \multirow{4}{*}{{Source model $\mathcal{S}$}} &VGG-16 & \bf 95.70 & 86.67 & 49.98 & 36.34 & 57.66  \\
		 & VGG-19 & 84.77& \bf 94.00 & 47.24 & 36.46 & 56.15 \\ 
		 & ResNet-18 & 76.49 & 72.18 & \bf 90.46 & 50.46 & 66.37  \\ 
		& ResNet-152 & 86.19 & 82.36 & 76.04 & \bf 90.40 & \bf 81.53  \\
		\bottomrule
	\end{tabular}
\end{table*}

\begin{table*}[t!]
    \caption{{\bf Transferability of our proposed non-targeted UAPs compared to other methods}. 
    The UAP is bounded by $L_{\infty}(\boldsymbol{r})\le \epsilon = 10$. Values of our method are taken from Table~\ref{transferability}. *Note that the results are reported from the respective paper. Best results are printed in \textbf{boldface}.}
    \centering
    \label{tab:comparison_transferability}
    \renewcommand{\arraystretch}{0.95} 
	\setlength\tabcolsep{7pt}
    \begin{subtable}{.5\linewidth}
     \centering
        \caption{Source model $\mathcal S$: VGG-16}
	    \begin{tabular}{c c c}
		\toprule
		Target Model $\mathcal{T}$ & Method  & Fooling Rate (\%) \\
		\midrule
		\multirow{5}{*}{VGG-19} & FFF*  & 41.98 \\
        & CIs*& 65.64 \\
        & UAP* & 73.10 \\
        & GAP & 79.14 \\
        & {\bf Ours} & {\bf 86.67} \\
		\midrule
		\multirow{5}{*}{ResNet-152} & FFF*  & 27.82 \\
        & CIs*  & 45.33 \\
        & UAP*  & \bf{63.40}\\
        & GAP  & 30.32 \\
        & {\bf Ours} & 36.34 \\
        \midrule
        ResNet-18
        & {\bf Ours} & \bf 49.98 \\
		\bottomrule
	    \end{tabular}
	\label{tab:comparison_transferabilitya}
    \end{subtable}%
    \begin{subtable}{.5\linewidth}
      \centering
        \caption{Source model $\mathcal S$: ResNet-152}
	    \begin{tabular}{c c c}
		\toprule
		Target Model $\mathcal{T}$ & Method  & Fooling Rate (\%) \\
		\midrule
		\multirow{5}{*}{VGG-16} & FFF* & 19.23 \\
        & CIs* & 47.21\\
        & UAP* & 47.00 \\
        & GAP & 70.45 \\
        & {\bf Ours} & {\bf 86.19} \\
		\midrule
		\multirow{5}{*}{VGG-19} & FFF*  & 17.15 \\
        & CIs* & 48.78 \\
        & UAP* & 45.50 \\
        & GAP & 70.38 \\
        & {\bf Ours} & {\bf 82.36} \\
        \midrule
        ResNet-18
        & {\bf Ours} & \bf 76.04 \\
		\bottomrule
    	\end{tabular}
    \label{tab:comparison_transferabilityb}
    \end{subtable} 
\end{table*}


\subsection{Comparison with Other Methods} \label{section.details55}
We compare our proposed approach in generating non-targeted UAPs with state-of-the-art methods in this field of research, i.e., fast feature fool (FFF) \cite{mopuri2018generalizable}, class impressions (CIs) \cite{reddy2018ask}, universal adversarial perturbation (UAP) \cite{moosavi2017universal}, and generative adversarial perturbation (GAP) \cite{poursaeed2018generative}. The results are shown in Table~\ref{tab:comparison}. In these experiments, the source model $\mathcal{S}$ and the target model $\mathcal{T}$ are the same.  Our proposed approach achieves a new state-of-the-art performance on {\it all} models on {\it both} $L_p$ norms, being on average 4\% absolute better in fooling rate with the $L_2$ norm, and even 11\% absolute better with the $L_{\infty}$ norm. Also, we compare the transferability of our produced UAPs to the same 
methods as before. The results for these experiments are shown in Table~\ref{tab:comparison_transferability}, where VGG-16 and ResNet-152 are used as the source model in Table~\ref{tab:comparison_transferabilitya} and Table~\ref{tab:comparison_transferabilityb}, respectively. It turns out to be advisable to choose a deep network as source model (ResNet-152), {\it{\rm since} our performance on the unseen VGG-16 and VGG-19 target models is about 12\% absolute better than earlier state of the art} ($L_{\infty}$ norm).

\section{Conclusions} \label{section.Conclusions}
In this paper, we have presented a new effective method to generate targeted and non-targeted universal adversarial perturbations (UAPs) in both white-box and black-box settings. Our proposed method shows new state-of-the-art fooling rate performance for non-targeted UAPs
on different classifiers. Additionally, our non-targeted UAPs show a significantly higher transferability across models when compared to other methods, given that we generated UAPs on the deepest network in the investigation. This is achieved by incorporating an additional loss term during training, which aims at increasing the activation of the first layer of the source model. Extending the proposed method to other tasks such as semantic segmentation will be subject of future research.


\medskip
\small

\bibliography{refs}

\end{document}